# Bootstrapped Adaptive Threshold Selection for Statistical Model Selection and Estimation

**Kristofer E. Bouchard,** Life Sciences and Computational Research Divisions, Lawrence Berkeley National Laboratory, Berkeley, CA

**Abstract**

A central goal of neuroscience is to understand how activity in the nervous system is related to features of the external world, or to features of the nervous system itself. A common approach is to model neural responses as a weighted combination of external features, or vice versa. The structure of the model weights can provide insight into neural representations. Often, neural input-output relationships are sparse, with only a few inputs contributing to the output. In part to account for such sparsity, structured regularizers are incorporated into model fitting optimization. However, by imposing priors, structured regularizers can make it difficult to interpret learned model parameters. Here, we investigate a simple, minimally structured model estimation method for accurate, unbiased estimation of sparse models based on Bootstrapped Adaptive Threshold Selection followed by ordinary least-squares refitting (BoATS). Through extensive numerical investigations, we show that this method often performs favorably compared to $L_1$ and $L_2$ regularizers. In particular, for a variety of model distributions and noise levels, BoATS more accurately recovers the parameters of sparse models, leading to more parsimonious explanations of outputs. Finally, we apply this method to the task of decoding human speech production from ECoG recordings.

## 1    Introduction

Many areas of neuroscience seek to understand neural activity by modeling its functional relationship with other features. Examples include estimating what the activity of a specific area encodes, as well as 'population' based methods, such as functional connectivity and population decoding[1]-[4]. These functional relationships, which often take the form of linear mappings, must be learned directly from the data. Examining the structure of the learned mappings can provide insight into neural computations[5]. Furthermore, optimizing model performance is important for decoding brain signals, such as for the functioning of brain-machine interfaces.

Neural responses, as well as the connectivity of the brain, often exhibit a high degree of sparsity[1],[6]. Because of this sparsity, even when data is abundant, model fitting of neuroscience data often result in under constrained optimizations, which can cause unstable results[7],[8]. One goal of regularization can be viewed as increasing the stability of statistical learning methods[7],[8]. Common approaches to regularization constrain the optimization problem by including a convex function of the magnitude of model weights, which imposes structure on the weight distribution[7],[8]. However, the imposition of this prior structure can hinder the interpretation of learned weights[8]. Furthermore, even with the inclusion of such structured regularizers, many model parameters are not set exactly to zero[9], [10], so in practice, post-hoc selection of model parameters is performed, usually outside of the model optimization[2].

In contrast to structured regularizers, pure 'variable selection' methods seek to only include parameters that contribute to model performance[11],[12]. One approach is to include a hard threshold operation in the optimization procedure, which sets weights that are below a threshold to zero, but this approach may suffer from high variability[12],[13]. Interestingly, thresholding has recently been proposed as an implementation for sparse coding in neural circuits[10]. Here, we propose a simple, efficient (one-dimensional meta-parameter) method for feature selection and estimation through adaptive threshold selection of model parameters followed by ordinary least-squares refitting. This method imposes minimal structure on the distribution of model weights, other than sparsity. Furthermore, by fitting the reduced model on training data, and selecting the threshold that optimizes these newly fit weights on predicting test data, we provide a principled means of jointly selecting and estimating features, alleviating the need for post-hoc selection often employed in neuroscience. On simulated data, this method provides more parsimonious models with comparable variability in model estimation and performance metrics. When applied to neural population decoding, it allows for accurate prediction of behavior.

## 2 Regularization in prediction and feature selection

We begin by discussing the role of regularization in the dual goals of statistical learning of prediction and variable selection for causal interpretation[8],[11]. We highlight a tension that exists in structured regularizers when the underlying sparse model diverges from the structure imposed by the regularizer [8],[14].

For expositional ease and direct connections to common neuroscience applications, we will restrict our discussion to the context of multi-linear regression with i.i.d. input values and zero-mean additive noise. Here, the output (e.g. neural response) $y \in \mathbb{R}$ is a weighted sum of the d-dimensional input features (e.g. pixel intensity), $x \in \mathbb{R}^d$, weighted by the vector $\beta \in \mathbb{R}^d$ and corrupted by a Gaussian noise process:

$$y = \sum_{j=1}^{d} \beta^j x^j + N(0, \sigma^2 I_{dxd}) \quad (1)$$

Given m measurements of input-output pairings $Z_i = (x_i, y_i), i = 1,\ldots m$, the goal is to find estimates of $\beta$ that minimize a convex loss function $\mathcal{L}(\beta, Z)$:

$$\hat{\beta} = \underset{\beta \in \mathbb{R}^d}{\operatorname{argmin}} \mathcal{L}(\beta, Z) \quad (2)$$

Here, we use the typical least-squares loss function:

$$\mathcal{L}(\beta, Z) = \sum_{i=1}^{m} (y_i - \beta x_i)^2 \quad (3)$$

We denote the estimate of the $i^{th}$ sample of $y$, given $\hat{\beta}$, as: $\hat{y}_i = \sum_{j=1}^{d} \hat{\beta}^j x_i^j$. Generally speaking, the curvature of the loss function is given by its Hessian, and, because the loss function (3) is convex, it has dimensions with positive curvature, and the magnitude of the gradient along a dimension describes its importance in loss reduction (Figure 1a).

In many problems of interest to neuroscience, the output y depends sparsely on x. In this case, only k dimensions of x are associated with $\beta \neq 0$. That is, $\|\beta\|_0 = k$ where $\|\beta\|_0$ is the $L_0$-norm of the vector: $\|\beta\|_0 = \sum_{j=1}^{d} 1(|\beta^j| > 0)$. If y depends sparsely on x, then, although there are dimensions of $\beta$ for which the loss function has positive curvature, there are dimensions for which the loss function is completely flat, and thus the gradient is 0 (e.g. $\beta_3$ in Figure 1a). In such cases, the optimization in (2) is under constrained, resulting in unstable solutions[7].

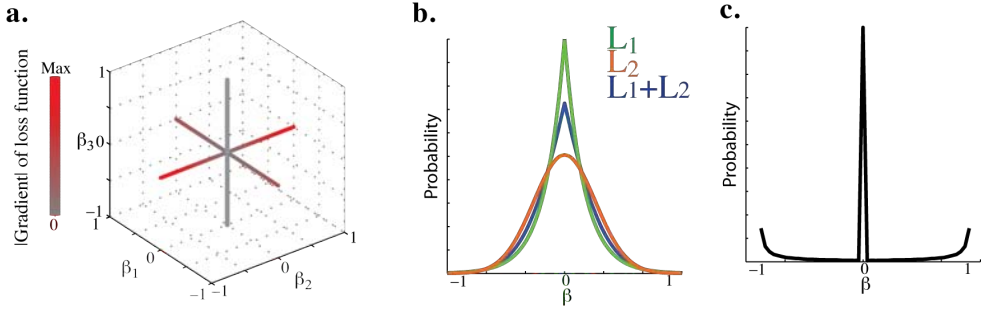

Figure 1. (a) Visualization of magnitudes of gradient of generic loss function. The goal of regularization is to discard dimensions with zero gradient (i.e. for which the loss function is flat, e.g. $\beta_3$), while minimally distorting the gradient along other dimensions. (b) Prior probability distribution imposed by three common structured regularization methods. (c) Example sparse model that diverges greatly from the prior distributions in (b).

Generally speaking, one goal of regularization is to exclude dimensions in which the gradient of the loss function is 0 (i.e. in which the loss function is flat), while minimally affecting the curvature of the loss function in other directions[7]. The typical approach to this problem is to provide additional constraints to the optimization in (2) by combining the empirical loss function (3) with a convex regularizer[7], [8], [12] $\mathcal{R}: \beta \to \mathbb{R}_+ 1$:

$$\hat{\beta} \in \underset{\beta \in \mathbb{R}^d}{\operatorname{argmin}} \{ \mathcal{L}(\beta, Z) + \lambda \mathcal{R}(\beta) \} \qquad (4)$$

Here, $\lambda$ is an empirically determined parameter that measures how strongly to weight the regularizer in the optimization.

Common choices for $\mathcal{R}$ are the $L_1$-norm ($\|\beta\|_1$) and $L_2$-norm ($\|\beta\|_2$) of the weight vectors, which correspond to the well-known LASSO[13], [15] and ridge regression algorithms, respectively. In particular, the $L_1$-norm, and the corresponding LASSO algorithm, have received much attention in sparse coding, as it is seen as the convex regularizer $\mathcal{R}: \beta \to \mathbb{R}_+$ closest to the $L_0$-norm (the natural norm to impose sparsity)[8],[13]. The $L_1$ and $L_2$ norms can be combined in the elastic net[16]:

$$\mathcal{L}(\beta, Z) = \frac{1}{2m} \sum_{i=1}^{m} (y - \beta x_i)^2 + \lambda_1 \|\beta\|_1 + \lambda_2 \|\beta\|_2 \qquad (5)$$

However, as is well known, these additional constraints impose prior expectations over the distribution of $\beta$. In particular, the $L_1$-norm ($\|\beta\|_1$) is equivalent to imposing a Laplace prior while the $L_2$-norm ($\|\beta\|_2^2$) imposes a Gaussian prior over the distribution of weights [13], [16]. The prior constraints on $\beta$ that arise from (5), and for which the other regularizers are special cases, can be approximated as:

$$\pi(\beta|\sigma^2) \propto \exp\left(-\frac{1}{2\sigma^2}(\lambda_1 \|\beta\|_1 + \lambda_2 \|\beta\|_2)\right) \qquad (6)$$

In Figure 1b, we display examples of $\pi(\beta|1)$ for three cases of $\lambda_1$ and $\lambda_2$, which exemplify the range of shapes and the centralizing nature of these priors.

When the goal of statistical learning is prediction, these additional constraints are critical in the case where the number of input dimensions (d) exceeds the number of samples (m) (d>>m) [7],[8]. If prediction accuracy is the only goal, as is often the case in machine learning, the values of the weights are irrelevant[8]. However, the priors associated with these constraints potentially come at a cost when selection and casual interpretation of features is the primary goal of statistical learning[8], as is often the case in neuroscience. This is because, to the degree to that the actual distribution $P(\beta \neq 0)$ diverges from the prior distribution $\pi(\beta \neq 0)$ [in the sense of, for example, $D_{KL}(\pi(\beta \neq 0)||P(\beta \neq 0))$], the priors will change the shape of the regularized loss function (equation 4) relative to the non-regularized loss function (equation 3) along dimensions with non-zero gradient (Figure 1a,b). Furthermore, when the magnitude of the (additive) noise ($\sigma^2$) is non-negligible, these structured regularizers often do not set many parameters exactly to 0 [9],[10].

For an example of the kind of model that diverges greatly from the prior given by equation 6, we imagine that y does indeed depend sparsely on x, in the sense that the support of the linear relationship between x and y (i.e. $\beta$) is small (that is, k<<d), but the actual distribution $P(\beta \neq 0)$ diverges greatly from the prior distribution $\pi(\beta \neq 0)$. The distribution in Figure 1c schematizes such an example. Here, not only is $\beta$ fairly sparse ($\frac{k}{d} = 1/3$), but there are a large number of weights at the extreme range of $\beta$. Such a weight distribution could correspond to neuronal tuning for "sharp" (i.e. high-derivative) edges.

## 3 Bootstrapped Adaptive Threshold Selection Algorithm

A good penalized estimation procedure for sparse models should satisfy three conditions: 1) estimate the model parameters while imposing minimal structure on $\beta \neq 0$ (unbiased), 2) set model parameters with 0 weight exactly equal to 0 (true sparsity); 3) provide reliable results for model estimates (low-variability)[14]. Here, we investigate the properties of a simple, efficient, minimally structured method for unbiased recovery of sparse models based on Bootstrapped Adaptive Threshold Selection (BoATS) followed by ordinary least-squares (OLS) refitting of the selected parameters. In this method, the hard threshold is some multiple of the expected value of the null-distribution of each model parameter, and is adaptively set to optimize model performance in terms of minimizing expected loss through cross-validation[11]. As with many approaches to parameter regularization/selection in statistical learning, our method finds its roots in the seminal work of Donoho and Johnstone, 1994 [12], and continued by many others (e.g. [9],[13]-[15],[17]).

In ATS-OLS refit, the first step is to estimate the null distribution of model weights ($\hat{\beta}_{null}$) by, for example, randomly permuting the relationship between inputs and outputs ($Z_{rnd}$) multiple times:

$$\hat{\beta}_{null} = E(\underset{\beta \in \mathbb{R}^d}{\mathrm{argmin}}\, \mathcal{L}(\beta, Z_{rnd})) \tag{7}$$

Then, divide the m measurements of input-output pairings $Z_i = (x_i, y_i), i = 1,...m$, into non-overlapping train ($Z_{trn}$), select ($Z_{slct}$), and test ($Z_{tst}$) sets for model training, selection, and testing (i.e. V-fold cross-validation). Next, derive an initial estimate of model parameters from $Z_{trn}$:

$$\beta_{init} = \underset{\beta \in \mathbb{R}^d}{\mathrm{argmin}}\, \mathcal{L}(\beta, Z_{trn}) \tag{8}$$

For a range of thresholds, set model parameters to zero if the magnitude of $\beta_{init}^j$ is less than a multiple ($\lambda_3$) of the expected value of the null magnitudes ($\hat{\beta}_{null}^j$):

$$if\ |\beta_{init}^j| < |\hat{\beta}_{null}^j| \times \lambda_3, then\ \beta_{sel}^j(\lambda_3) = 0 \tag{9}$$

and re-fit the d-n $\beta_{sel}^j \neq 0$ on $Z_{trn}$ using ordinary least-squares:

$$\beta_{sel}(\lambda_3) = \underset{\beta \in \mathbb{R}^{d-n}}{\mathrm{argmin}}\, \mathcal{L}(\beta, Z_{trn}) \tag{10}$$

Finally, select the optimal model parameters ($\beta_{opt}$) as the $\beta_{sel}$ that minimize the expected loss on out-of-sample data ($Z_{sel}$) across thresholds ($\lambda_3$):

$$\beta_{opt} = \underset{\lambda_3}{\mathrm{argmin}}\, \mathcal{L}(\beta_{sel}(\lambda_3), Z_{sel}) \tag{11}$$

We note that other reasonable choices for $\hat{\beta}_{null}$ could have been made, e.g. $\hat{\beta}_{null} = \frac{1}{d}(|E[y]| + E[y^2] - E[y]^2)$. We calculated expected predictive performance ($R^2$) and model parsimony (BIC) on data ($Z_{tst}$) not used in parameter training or selection: $R^2 = 1 - \frac{\sum(y - \beta_{Opt}x)^2}{\sum(y - E[y])^2}$; BIC = $T \ln(\frac{1}{T-1}\sum(y - \beta_{Opt}x)^2) + k \ln(T)$, where k = $\|\beta_{opt}\|_0$, and $T$ is the number of samples in ($Z_{tst}$).

## 4. Numerical investigations

We compared the performance of the BoATS procedure to several well known structured regularizers. The same cross-validation procedure ($Z_{trn} = \frac{4 \times m}{5}$), select ($Z_{slct} = \frac{m}{10}$), and test ($Z_{tst} = \frac{m}{10}$) described in Section 3 was used to fit and evaluate all estimation methods. All models had k = 100 parameters β > 0. We varied the sparsity by increasing or decreasing the number of parameters with β = 0. We varied the number of samples (m) as a multiple of the total number of model dimensions. To arrive at estimates of expected values as well as variability of parameter estimates and performance metrics, the cross-validation algorithm was included in a 100-iteration bootstrap procedure, in which the cross-validation sets were randomly selected on each iteration. For optimizing regularization parameters for the structured regularizers, first a broad sweep across several orders of magnitude was performed. From this initial sweep, a range of parameters around the (cross-validated) optimal value was selected for a more fine-grained sweep to ensure that true optimal values were selected. The BoATS method had only a single parameter sweep. All code was written in MATLAB and used built-in functions ('regress', 'ridge', 'lasso'). The elastic net used the 'lasso' function with a 50/50 split between the $L_1$ and $L_2$ norms. The expected values of optimal model parameters ($\beta_{opt}$) were taken as the mean model parameters associated with meta-parameter ($\lambda_1$, $\lambda_2$, or $\lambda_3$) that gave best mean out-of-sample predictive performance, where the means are taken across bootstrap samples.

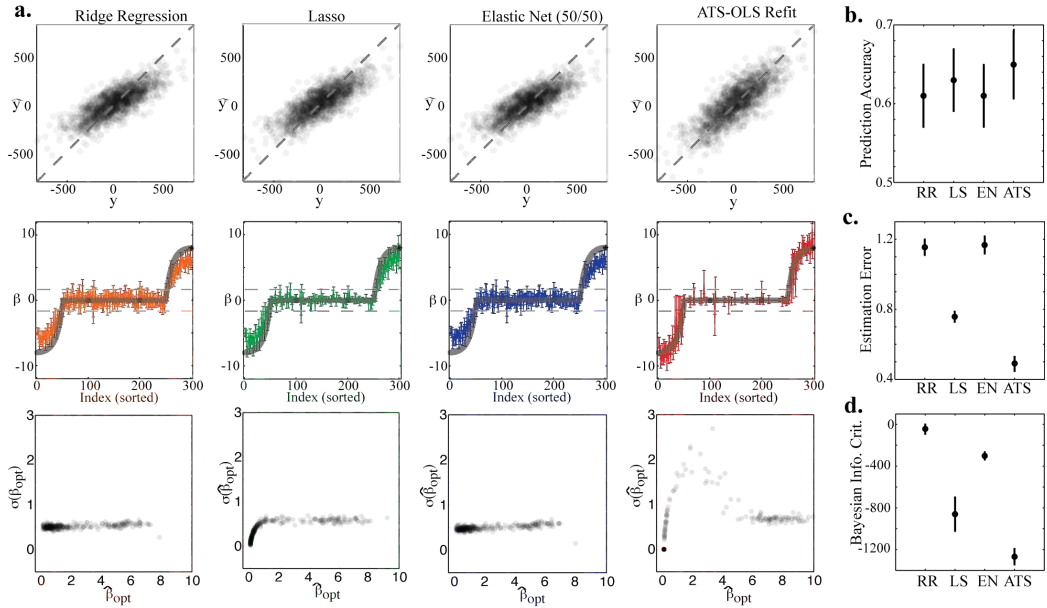

Figure 2. (a) Comparison of Ridge Regression, LASSO, Elastic Net (50/50), and Boats methods. (top) Predicted values of model output (ŷ) vs. actual values (y) based on optimal weights estimated with different methods. (middle) Value of each model parameter (transparent black) and mean ± s.d. of optimal estimated values. Dashed grey horizontal lines demarcate the mean ± s.d. of $|\hat{\beta}_{null}|$. (bottom) Scatter plot of variability in parameter estimation as a function of mean value. (b-d) Quantification of performance of different methods. (b) Prediction Accuracy ($R^2$), (c) Estimation Accuracy (RMS($\beta, \hat{\beta}_{opt}$)), (d) Model Parsimony (Bayesian Information Criterion). (RR: Ridge Regression, LS: LASSO; EN: Elastic Net, ATS: Adaptive Threshold Selection-OLS). Results are presented as mean ± s.d. across bootstrap samples.

## 4.1 Example

The results in Figure 2 compare the performance of different estimation methods when the underlying model distribution is of the form schematized in Figure 1c, with 200 null-dimensions (Sparsity = $1 - \frac{k}{d} = \frac{2}{3}$) and a sample size ratio $\frac{m}{d}$ = 5 (m = 1500). The top two panels in Figure 2a show the effect of the centralizing priors (equation 6) on the estimation of this model: the model estimates from the structured regularizers are biased towards smaller values ('shrinkage'). This can be seen as the narrower range of ŷ compared to y in the top panel of Figure 2a, and is directly observed in the middle panel Figure 2a. This bias results from the large regularization parameter required to identify the large number of null-dimensions in this model. In contrast, BoATS does not have this bias, on average estimating model parameters with both large and intermediate weights well (middle panel of Fig.2a), resulting in smaller average estimation error of model parameters (RMS($\beta_{opt}, \beta$), Fig.2c). Additionally, relative to the structured regularizers, the BoATS approach generally results in very low variability estimates of the zero-weights (null dimensions) and comparable variability of large weights, while weights near the threshold had increased variability because of noise (Fig.2a bottom). Finally, the structured regularizers (including LASSO) do not correctly identify all null dimensions (Fig.2a, middle panel), leaving many with non-zero weights. This can also be understood in terms of the tension created by having a large number of null dimensions and many large magnitude weights, as the trade-off between estimating these dimensions must be balanced by the structured regularizers. In contrast, BoATS correctly identifies almost all null-dimensions, setting them exactly to zero (Fig.2a). This results in a more parsimonious model for BoATS compared to the other methods, explaining more output variability with smaller model support (Fig.2b,d).

### 4.2    Performance across sparsity and sample size

We next compared the performance of these methods at estimating the parameters of this class of model (Fig.1c) while varying both the sparsity and the number of measurements, relative to the dimensionality of the model (Sparsity $= 1-\frac{k}{d}$, Samples $=\frac{m}{d}$). The colored surfaces in Figure 3 display the estimation error (Fig.3a: RMS($\beta_{opt}, \beta$)), estimation variability (Fig.3b: $\frac{1}{d}\sum_{j=1}^{d} \sigma(\beta_{Opt}^{j})$), and output prediction accuracy (Fig.3c: $R^2$). As expected, across methods, estimation results were very poor when the number of measurements was approximately equal to or only slightly greater than the number of model dimensions (here, $\frac{m}{d} < 3$), reiterating the difficulty of interpreting model parameters with small numbers of measurements (Fig.3a,c). Furthermore, in the regime d~m, the BoATS algorithm performed considerably worse then its more structured counterparts, yielding highly variable model estimates (Fig.3b) and lower prediction accuracy (Fig.3c). This emphasizes the importance of structured regularizers when attempting to accurately predict with few measurements[7]. However, in the regime when sparse models could be accurately predicted (here, $\frac{m}{d} \geq 3, 1-\frac{k}{d} = 0.5$), we found that the BoATS method generally performed favorably compared to the other methods, resulting in smaller model estimation error (Fig.3a), with comparable amounts of variability in parameter estimates (Fig.3b) and slightly better output prediction accuracy (Fig.3c).

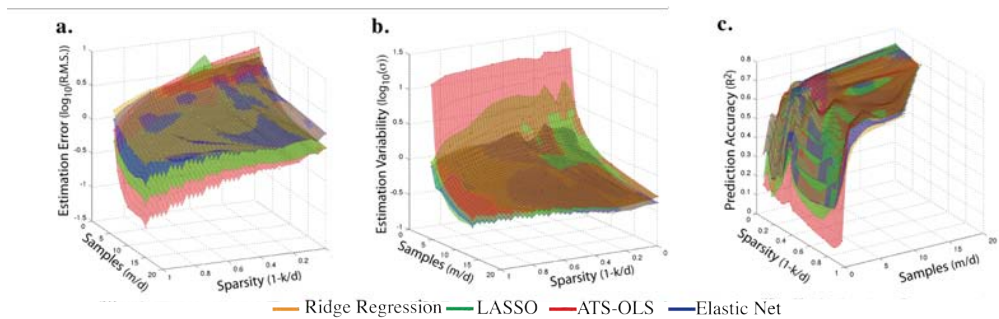

Figure 3. Comparison of Ridge Regression, LASSO, Elastic Net (50/50), and BoATS methods for different degrees of relative sparsity and number of measurement samples.
(a) Estimation error; (b) Estimation variability; (c) Prediction accuracy. Each colored surface corresponds to mean results across bootstrap samples for models estimated by different methods.

### 4.3    Performance across different model distributions
We observed similar results across a range of model distributions. In Figure 4, we present performance metrics across four different model distributions for a range of sparsities (for the case $\frac{m}{d} = 3$, which is where prediction accuracy stabilized across all estimation methods). The histograms in Figure 4a schematize the model distributions (black histograms), while the grey bar at zero with bidirectional arrow demarcates that we manipulated sparsity by simply changing the number of model parameters equal to zero. Generally speaking, across model distributions, BoATS resulted in smaller estimation error compared to other methods as the sparsity of the model increased (Fig.4b), with slightly greater or comparable estimation variability (Fig.4c) relative to Lasso.

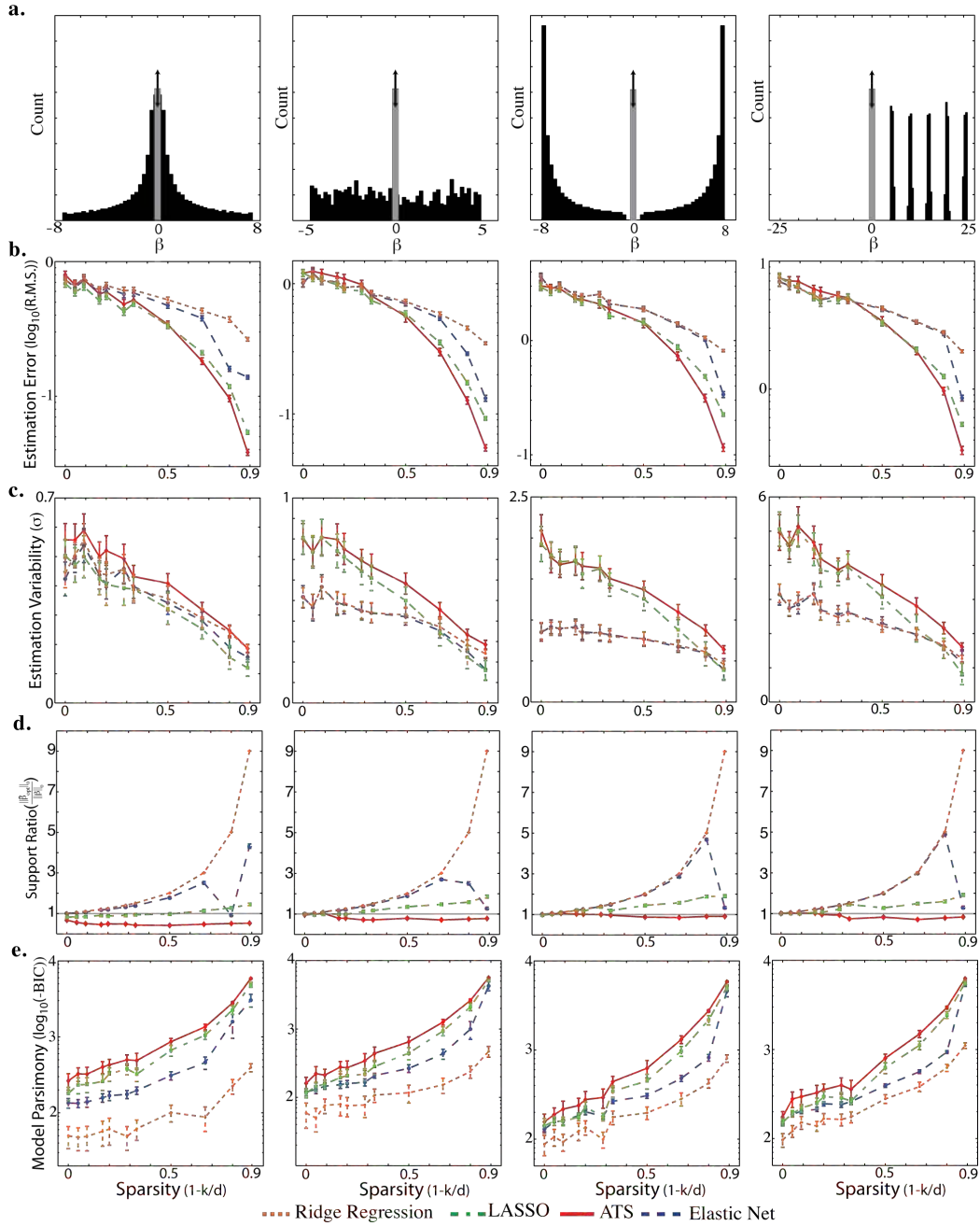

Figure 4. Comparison of Ridge Regression, LASSO, Elastic Net (50/50), and BoATS methods for different model distributions with varying degrees of relative sparsity. (a) Schematics of different distributions used. Far left: Laplace distribution; center left: uniform distribution; center right: symmetric increasing exponential; far right; asymmetric clustered.

(b)Estimation error; (c) Estimation variability; (d) Support ratio; (e) Model Parsimony. Results are presented as mean ± s.d. across bootstrap samples.

Across all models and sparsities, BoATS selected smaller models relative to the other methods, and actually selected fewer parameters than were in the actual model (Fig.4d). This was most pronounced for the Laplace distribution (Fig.4, far-left), while for the other distributions BoATS resulted in models with support that was very close to the support of the actual model. Together with the comparable or superior prediction performance, the small models estimated by BoATS resulted in more parsimonious predictors than the other methods (Fig.4e). This suggests that the model parameters that are being selected/estimated by BoATS are those that are contributing to estimation accuracy above noise levels, while (e.g.) LASSO retains many small parameters that cannot be well estimated given the noise levels[9]. As suggested by Figure 3, increasing samples reduced model estimation error for BoATS relative to other methods, equalized variability, and maintained the superiority of BoATS in terms of model parsimony.

### 4.4 Comparison across noise magnitudes

The results presented above were for linear input-output relationships (equation 1) with a modest amount of additive white noise ($\sigma^2 = 0.2 \times \sum |\beta|$). We next compared the performance of model estimation for BoATS to other methods while varying the noise magnitude. Specifically, we varied the standard deviation of the additive white noise as a multiple of the summed weight magnitudes ($\sigma^2 = c \times \sum |\beta|$). The plots in Figure 5 show results for the model distribution displayed on the far right of Figure 4a for six values of the multiplicative factor c (Sparsity = 0.66).

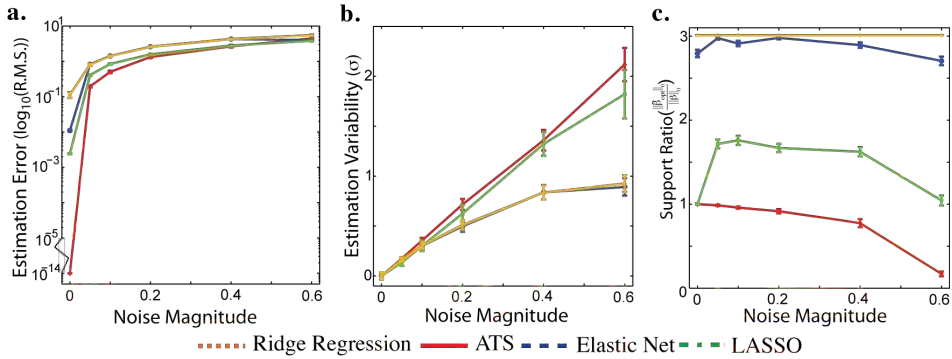

Figure 5. Comparison of Ridge Regression, LASSO, Elastic Net (50/50), and BoATS methods for different magnitudes of additive white noise. Results are presented as mean ± s.d. across bootstrap samples.

As expected, both the error in model estimation and variability in estimated model parameters increased for all methods as the magnitude of the noise increased (Fig.5a,b). The support for models estimated by ridge regression and elastic were much larger than the actual model, models estimated by lasso had support that were also larger than target for all but lowest and highest noise regimes, while the support for models estimated by BoATS were smaller than the target, and decreased monotonically with increasing noise levels. Strikingly, the decrease in estimation error at lower noise magnitudes was more pronounced for BoATS than the other methods. In the noiseless case, this resulted in estimation errors that were more than 11 orders of magnitude less than estimation errors from the other methods (Fig.5a). BoATS resulted in model estimates that were generally more accurate, comparably variable, with better parameter selection (relative to the true model) than the other methods tested, for all but the nosiest conditions examined here.

## 5 Application of ATS-OLS Refit to neuroscience data

To demonstrate applicability to neuroscience data, we applied the BoATS refit method to the problem of population decoding of behavior. For this exercise, we used previously described data recorded with ECoG from the human speech sensory-motor cortex[18]. Specifically, we used BoATS refit to train linear models to predict an acoustic feature ($F_2/F_1$) of produced vowels (/a/, /i/, and /u/) from the amplitude of high-frequency signal extracted from field potentials on a single-trial basis. We first trained a decoder to predict the acoustics across these vowels, and found that a linear model estimated by BoATS could predict ~80% of this variability ($R^2 = 0.81$). However, this decoder did not predict the within vowel variability (shallow slopes of colored lines of fits within a vowel). We thus trained separate linear models for each vowel individually, and found that even this within vowel variability can be predicted from the ECoG recordings (upwards of 30%, e.g. Figure 6b). These results demonstrate that BoATS can be used to neuroscience data.

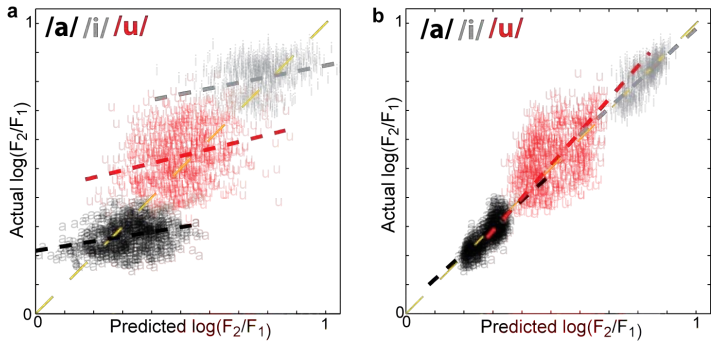

Figure 6. Results of applying BoATS to the task of decoding produced human speech from ECoG recordings. a. Accurate decoding of across vowel acoustics. b. Accurate decoding of within vowel acoustics. m = 2502 total vocalizations.

## 6  Discussion

We have shown that adaptive threshold selection with ordinary least squares refitting results in more parsimonious models, improving output predictions with fewer parameters, while only modestly increasing variability. Comparison to other methods for model selection, such as the BoLASSO, is an important future direction. As pointed out by[14], neither a hard threshold selection method or any $L_q$-norm regularization simultaneously satisfy all conditions of an idealized regularized estimator: while unbiased, hard thresholds have increased variability due to the discrete nature of selection coupled with noise, and conversely, $L_q$-norm regularizers, while giving low-variability estimates because of the smooth nature of the regularizer, are biased towards smaller weights. The smoothly clipped absolute deviation (SCAD,[14]) and the relaxed Lasso[9] both attempt to resolve this by introducing a two-parameter regularization function. However, the systematic search of the 2-dimensional parameter space associated with the penalties makes them rarely used in practice, and so we restricted comparisons to methods commonly used on neuroscience data.

Our approach is easily extended to other output/input distributions, to mixed continuous and discrete data, and to non-linear models. Additionally, calculation of information theoretic limits of minimax estimation properties, investigation of non-orthogonal designs, and application to estimating receptive fields of sensory neurons are worthwhile.

**Acknowledgments:** We thank Edward Chang for utilization of human ECoG recordings.